\documentclass[journal]{IEEEtran}

\usepackage{cite}
\usepackage{amsmath,amssymb}
\usepackage{graphicx}
\usepackage{array}             
\usepackage{booktabs}
\usepackage{multirow}
\usepackage{pifont}            
\usepackage[hidelinks]{hyperref}
\usepackage{url}

\newcommand{\NItems}{201}
\newcommand{\NFamilies}{18}
\newcommand{\NSources}{9}

\newcommand{\Chance}{25.0}
\newcommand{\AudioMinutes}{23}
\newcommand{\AudioMean}{6.8}
\newcommand{\AudioMedian}{3.5}
\newcommand{\AudioMin}{0.1}
\newcommand{\AudioMax}{48}
\newcommand{\NFourWay}{201}

\newcommand{\NSynth}{48}
\newcommand{\NAbsolute}{95}
\newcommand{\NCategorical}{60}
\newcommand{\NComparative}{46}

\newcommand{\NRejected}{11}
\newcommand{\NPlanned}{9}
\newcommand{\NGoldAgree}{93}
\newcommand{\NGoldDisagree}{0}
\newcommand{\NGoldNA}{108}
\newcommand{\NChkRuleCovered}{189}
\newcommand{\NChkRuleMissing}{12}
\newcommand{\NChkRuleAgree}{129}
\newcommand{\NChkRuleDisagree}{0}
\newcommand{\NChkRuleNA}{60}

\newcommand{\NChkRemeasAgree}{79}
\newcommand{\NChkRemeasDisagree}{18}
\newcommand{\NChkRemeasNA}{92}

\newcommand{\NChkToolAgree}{43}
\newcommand{\NChkToolDisagree}{5}
\newcommand{\NChkToolNA}{141}
\newcommand{\MAName}{Gemini\nobreakdash-3.8\nobreakdash-flash}
\newcommand{\NMAFullWrong}{3}
\newcommand{\NMABaseRuleReversal}{16}
\newcommand{\NMAAnyReversal}{18}
\newcommand{\MANone}{29.3}

\newcommand{\MAKnowledge}{31.3}

\newcommand{\MAPercept}{98.5}

\newcommand{\MANoaudio}{23.9}

\newcommand{\BtNMACapable}{59}

\newcommand{\BtNMAKnowledge}{20}

\newcommand{\BtNMAPerception}{122}

\newcommand{\NProbeEquivalent}{10}
\newcommand{\ProbeGainEquivalent}{+0.0}
\newcommand{\NProbePartial}{5}
\newcommand{\ProbeGainPartial}{+12.1}
\newcommand{\NProbeWrong}{3}
\newcommand{\ProbeGainWrong}{-8.3}

\newcommand{\QAbsoluteN}{95}
\newcommand{\QAbsolutePercept}{64}

\newcommand{\QComparativeN}{46}
\newcommand{\QComparativePercept}{70}

\newcommand{\QCategoricalN}{60}

\newcommand{\QCategoricalCapable}{45}

\begin{document}

\title{AuDiR: Diagnosing Audio Reasoning\\with Rule and Percept Interventions}

\author{Chengwei~Wei,
        Yi~Ma,
        and~Nancy~F.~Chen,~\IEEEmembership{Senior Member,~IEEE}
\thanks{C. Wei and N. F. Chen are with the Institute of Advanced Intelligence and Computing
(IAIC) and the Centre for Frontier AI Research (CFAR), Agency for Science, Technology and
Research (A*STAR), Singapore.}
\thanks{Y. Ma is with the Institute of Advanced Intelligence and Computing (IAIC), Agency for
Science, Technology and Research (A*STAR), Singapore.}
\thanks{Corresponding author: Chengwei Wei (e-mail: chengwei7272@gmail.com).}
}

\markboth{AuDiR: Audio Diagnostic Reasoning}%
{Wei \MakeLowercase{\textit{et al.}}: AuDiR}
\maketitle

\begin{abstract}
Answer accuracy alone does not distinguish failures in acoustic perception, rule knowledge,
and rule application. AuDiR (Audio Diagnostic Reasoning) examines these distinctions through
controlled input interventions. Each item contains an audio clip, a multiple-choice question,
a task-general rule, and a textual description of the acoustic observations needed to answer.
Three conditions progressively supply the rule and observations; a fourth removes the audio
while retaining the question and rule. The first correct diagnostic condition assigns an
operational response label, without identifying a unique internal cause. The current benchmark
contains \NItems{} items in \NFamilies{} families drawn from \NSources{} sources, with answer
references documented separately from percept extraction. In a preliminary evaluation of
\MAName{}, accuracy is \MANone{}\% without supplied information, \MAKnowledge{}\% with the rule,
and \MAPercept{}\% with the observations also supplied. No-audio accuracy is \MANoaudio{}\%,
with a confidence interval containing the \Chance{}\% chance rate. A family-level knowledge
probe associates incomplete rule recall with larger gains from rule provision. These results
show the benefit of supplying acoustic observations as text, while incomplete validation,
possible recognition shortcuts, and non-monotonic responses limit stronger failure-attribution
claims.

\end{abstract}


\IEEEpeerreviewmaketitle

\section{Introduction}\label{sec:intro}
\IEEEPARstart{A}{udio} reasoning benchmarks such as MMAR~\cite{mmar}, MMAU~\cite{mmau} and
STAR-Bench~\cite{starbench} measure whether a model answers correctly. Accuracy alone does not distinguish among
three possible sources of error:

\begin{itemize}
  \item \textbf{Missing knowledge.} The model did not know the rule the task needs: that a
        passing car's tone shift gives its speed, or that a shaft tone at $f$\,Hz means $60f$\,RPM.
  \item \textbf{Failed perception.} The model knew the rule but did not recover the quantity the
        rule consumes from the audio.
  \item \textbf{Failed application.} The model had the rule and the quantity and still reached the
        wrong answer.
\end{itemize}

Answer scores alone cannot separate these cases. Post-hoc analysis can provide additional
information: MMAR manually categorizes a sample of failed predictions as perceptual, knowledge,
reasoning, or other errors. Such analysis requires annotation and relies on the model's written
explanation, which need not reflect the computation that produced its answer.

AuDiR (Audio Diagnostic Reasoning) makes the attribution part of the item instead. We write
each item as three separable parts: a rule that can be stated in language, an acoustic observation to be recovered from the clip, and a question that needs the rule applied to the percept. We then run
the item under three diagnostic conditions: audio and question alone, then with the rule,
then with the percept text as well. The first correct condition assigns a response-pattern
label associated with capability, knowledge or perception; failure under all three is assigned
to application. The interpretation depends on the supplied information being sufficient and
correct, and does not assume that responses improve monotonically. A fourth, separate
condition removes the audio to probe answerability from the question and rule alone.

The protocol records responses to four controlled inputs for every item. This reduces the need
for post-hoc annotation of model explanations, although constructing and validating the items
still requires human work.

We contribute:
\begin{itemize}
  \item An item design and operational diagnostic protocol: rule, percept, and question as
        separate fields, evaluated under three diagnostic conditions and one control
        (Section~\ref{sec:design}).
  \item A construction procedure that separates percept extraction from answer references and
        documents rejected candidate pairings (Section~\ref{sec:pipeline}).
  \item A benchmark build with explicit provenance and contamination flags, and no-audio
        responses reported in aggregate (Sections~\ref{sec:structure} and~\ref{sec:validation}).
  \item A preliminary evaluation of \MAName{} over all \NItems{} items, including intervention
        response profiles and a family-level knowledge-recall probe (Sections~\ref{sec:setup}
        and~\ref{sec:analysis}).
\end{itemize}

\section{Related Work}\label{sec:related}

\subsection{Audio reasoning benchmarks}
Benchmarks for large audio language models have grown quickly in size and in the range of skills
they cover. AIR-Bench~\cite{airbench} pairs roughly 19k single-choice questions over nineteen
foundation tasks with 2k open-ended items. MMAU~\cite{mmau} assembles 10k clips over speech,
environmental sound and music, annotated for 27 skills. MMAR~\cite{mmar} takes the opposite trade
and curates 1{,}000 audio--question--answer triplets from real videos, filtered through repeated
correction passes, with a deliberate tail of items needing graduate-level domain knowledge.
STAR-Bench~\cite{starbench} narrows the target to spatio-temporal and physical reasoning and, like
us, uses synthesized and physics-simulated audio so that the governing quantity is known exactly.

These benchmarks differ in scope, but their answer scores do not directly separate the three
failure modes considered here.

Audio necessity is better served than failure attribution, though only in aggregate. MMAU replaces
the audio with Gaussian noise and reports that some models are barely affected. MMAR runs the same
substitution and finds that every model degrades, while noting that one still scores above chance
on noise alone, which it attributes to a residual language prior that survived a three-stage
review. STAR-Bench substitutes machine-generated captions for the audio and reports that
caption-only answering costs MMAU only 5.9\% and MMAR 9.0\%, against 31.5\% and 35.2\% on its own
items, and uses the contrast to argue that earlier benchmarks test what text can already carry.
All three are model-level diagnostics run on a finished benchmark. None is an admission test: an
item that survives the substitution is reported, not removed. AuDiR includes an explicit
no-audio condition in its construction framework. The current evaluation reports this control
in aggregate; implementing a validated per-item admission test remains further work.

\subsection{Stage-level failure attribution}
Three routes to a stage-level diagnosis exist, and all three are applied after the benchmark is
built.

The first reads the model's output. MMAR~\cite{mmar} has annotators read 100 failed predictions
from one model and label each as a perceptual, knowledge, reasoning or other error.
This provides useful error categories, but requires manual annotation and interprets written
explanations rather than observing internal computation.

The second reads aggregate score patterns. STAR-Bench~\cite{starbench} evaluates nineteen models
and concludes that proprietary systems are bottlenecked by fine-grained perception while open
models lag across perception, knowledge and reasoning together. The vocabulary is close to ours and
the evidence is not per item: the conclusion is inferred from how model averages move across task
families, so it describes a model on a benchmark rather than a model on an item, and it cannot say
which items would be fixed by which intervention.

The third opens the model. SonicBench~\cite{sonicbench}, concurrent with this work, isolates twelve
low-level physical attributes such as pitch, loudness and spatial location, synthesizes stimuli in
which each is controlled, and asks binary questions about it. Most audio language models sit near
the 0.50 chance rate where human listeners reach 0.91, and explicit reasoning adds little.
SonicBench then localizes the loss by probing: a two-layer probe on the frozen final encoder
output, trained per attribute, reaches 60\% or better even where the end-to-end model answers at
chance, so the cue is present in the representation and the loss occurs after the encoder.

Representation probing and input interventions provide different evidence. Probing tests whether
an acoustic attribute can be decoded from model representations and requires access to those
representations. AuDiR instead tests whether supplying a rule or observation changes the answer,
so it can also be applied to API-only models. A correct answer after supplying percept text does
not identify where information was lost within the model.

Caption substitution and AuDiR's percept condition also differ. A caption summarizes a clip;
our percept text states the observations selected for a particular question. The separate
no-audio condition retains neither the clip nor these observations. It probes text-only
answerability in aggregate, without certifying individual items as shortcut-free.
The protocol is limited to tasks whose required observations can be expressed in text.

\subsection{Diagnostic datasets and input ablation}
Building a dataset so that its errors are interpretable has a long precedent outside audio.
CLEVR~\cite{clevr} generates scenes and questions from a known grammar, so a failure can be traced
to a specific reasoning operation. HANS~\cite{hans} constructs inference cases in which a named
heuristic gives the wrong answer, making reliance on it visible in the score.
CheckList~\cite{checklist} tests capabilities separately instead of reporting one aggregate. AuDiR
follows this line with a different controlled variable: not the structure of the input, but how much
of the solution is handed over.

Our fourth condition belongs to a second established family. Hypothesis-only baselines showed that
a large share of natural language inference items could be classified without reading the
premise~\cite{hyponly, artifacts}, and the same logic motivated rebalancing visual question
answering so that answers could not be recovered from the question's language prior
alone~\cite{vqamatter, vqacp}. Removing the input the task is nominally about, and observing that
performance survives, is what exposed the shortcut. AuDiR combines this control with explicit
rule and percept interventions. The control probes text-only answerability, while the
diagnostic conditions describe how responses change when information is supplied.

Table~\ref{tab:related} places AuDiR against the audio benchmarks above on the two axes that matter
here. The systems all of this targets are audio-conditioned language
models~\cite{ltu, salmonn, audioflamingo, qwen2audio, qwen25omni}; Section~\ref{sec:analysis}
reports the preliminary evaluation of one model.

\begin{table*}[t]
\centering
\caption{AuDiR against existing audio reasoning benchmarks. ``Failure attribution'' asks whether
the benchmark says \emph{which} step failed, and on what evidence. ``Audio-necessity check'' asks
whether items are tested for being answerable without the audio, and what is done with those that
are. Entries reflect what each paper reports; a dash means the paper reports none.}
\label{tab:related}
\small
\begin{tabular}{@{}>{\raggedright\arraybackslash}p{1.80cm}>{\raggedright\arraybackslash}p{1.10cm}>{\raggedright\arraybackslash}p{4.50cm}>{\raggedright\arraybackslash}p{4.95cm}>{\raggedright\arraybackslash}p{3.95cm}@{}}
\toprule
Benchmark & Items & Failure attribution & Audio-necessity check & Answer ground truth \\
\midrule
AIR-Bench~\cite{airbench} & $\sim$21k & --- & --- & human annotation, model-judged \\
MMAU~\cite{mmau} & 10k & --- & noise substitution, per model, reported & human annotation \\
MMAR~\cite{mmar} & 1{,}000 & post-hoc, by reading 100 failed predictions from one model & noise substitution, per model, reported & curated human annotation \\
STAR-Bench~\cite{starbench} & 2{,}353 & from aggregate score patterns across task families & caption substitution, per benchmark, used as motivation & procedural synthesis and simulation \\
\midrule
\textbf{AuDiR} (ours) & \NItems{} & per-item intervention response patterns & no-audio accuracy, reported in aggregate & independent-reference design; provenance audit pending \\
\bottomrule
\end{tabular}
\end{table*}

\section{The AuDiR Benchmark}\label{sec:design}

\subsection{Overview}
AuDiR evaluates audio reasoning through controlled changes to the information supplied to a
model. Each multiple-choice item combines an audio clip and a question with a task-general
rule and a textual representation of the relevant acoustic observations. Three diagnostic
conditions progressively supply the rule and the observations; a fourth condition removes
the audio to probe text-only answerability. The first successful diagnostic condition defines
an \emph{operational diagnosis}: a label for the observed response pattern, whose interpretation
depends on the quality of the supplied information.

The current build contains \NItems{} items in \NFamilies{} families drawn from \NSources{}
sources. This section defines the item components, the input conditions, the assignment rule and
the record format of an item. Section~\ref{sec:sources} describes the sources, the composition
they yield and the construction procedure; Section~\ref{sec:setup} reports the baseline
evaluation.

\subsection{Knowledge, percept, and question}
The item design separates the observation to be recovered from the rule that consumes it.
Let $A$ denote the audio and $Q$ the question, its answer choices and any stated constants.
The three components have distinct roles:

\begin{itemize}
  \item \textbf{Knowledge} ($K$) states a rule shared across a task family. It may include
        general constants, but not the item's acoustic observations or target answer.
  \item \textbf{Percept} ($P$) states the item-specific observations that a listener would
        need to recover from $A$, such as two frequencies or an onset and an offset. This
        text is withheld in the base condition.
  \item \textbf{Question} ($Q$) asks for a conclusion obtained by applying $K$ to those
        observations. Item-specific non-acoustic givens are included here and kept fixed
        across conditions.
\end{itemize}

\begin{figure*}[t]
\centering
\includegraphics[width=\textwidth]{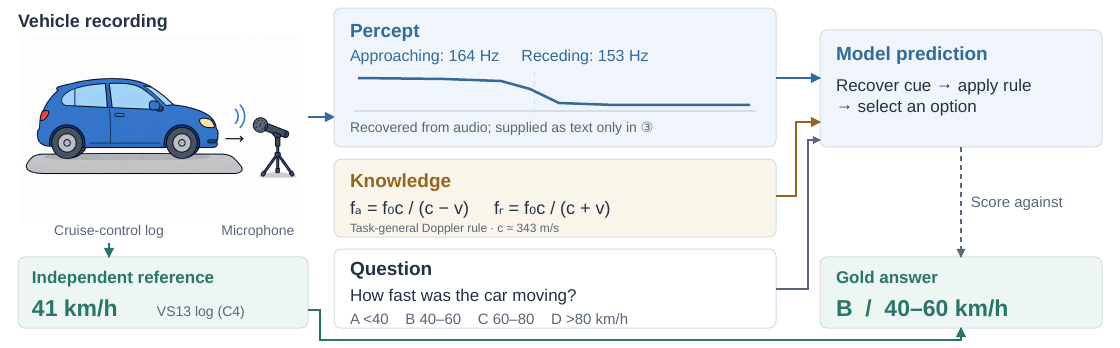}
\caption{An AuDiR item illustrated with vehicle speed. The model must recover the percept
from the audio and apply the task-general rule to answer the question. The tone trajectory
and frequencies are illustrative explanations, not text supplied in the base condition.
The gold option is obtained separately by binning the cruise-control speed VS13 records
for that pass; it is not estimated from the audio. This separation implements criterion~C4
(Section~\ref{sec:pipeline}).}
\label{fig:item}
\end{figure*}

\emph{Example (vehicle speed).} The audio is a car passing a roadside microphone at constant
speed. The percept is the pair of engine-tone frequencies heard while the car approaches and
recedes. The question is ``how fast was the car moving?'' The knowledge
is the Doppler relation, $f_a = f_0 c / (c - v)$ while approaching and $f_r = f_0 c / (c + v)$
while receding, with $c \approx 343$\,m/s. Applying this relation to the two frequencies gives
a speed estimate, which is mapped to an answer bin. The \emph{gold} bin is determined
separately from the speed the driver held on cruise control during that pass, which VS13
records alongside the audio, rather than from an audio-based speed estimate.
Fig.~\ref{fig:item} illustrates the audio-based prediction path and the independent
measurement path used to obtain the gold answer.

Three requirements govern the intended decomposition. First, $K$ must be task-general and
free of item-specific observations and answers. Second, $P$, together with the givens in $Q$,
must supply all observations needed to apply $K$. Third, $P$ must preserve the claimed
application step: it should state observations rather than the target answer. These are
requirements to audit, not guarantees provided by the field names. In particular, an event-rate
percept must include duration as well as count, and a category question must not receive its
answer directly in the percept field (Section~\ref{sec:diagnosis}).

Ground-truth provenance is a separate requirement. The gold answer should come from an
independent measurement or a quantity fixed by construction, rather than from the same
audio estimator used to write $P$. An externally supplied annotation is not by itself proof
of this independence; its derivation must also be documented (Section~\ref{sec:pipeline}).

\subsection{The four conditions}
Table~\ref{tab:conditions} specifies the complete inputs. Each condition is a separate
inference run, with the same question, choices and scoring rule. Conditions~\ding{172}--\ding{174}
add information cumulatively; condition~\ding{175} is a separate control, not a further
step in that sequence. In the vehicle example, \ding{173} adds the Doppler rule,
\ding{174} also states the two frequencies, and \ding{175} retains the question and rule
but supplies neither audio nor percept text.

\begin{table}[t]
\centering
\caption{Complete inputs for the four conditions. $Q$ includes the question, choices and
fixed givens; $P$ is the textual representation of the acoustic observations.}
\label{tab:conditions}
\small
\setlength{\tabcolsep}{3pt}
\begin{tabular}{@{}lll@{}}
\toprule
Condition & Full input & Intervention \\
\midrule
\ding{172} Base         & $A+Q$       & No supplied rule or percept \\
\ding{173} $+$Knowledge & $A+Q+K$     & Rule supplied \\
\ding{174} $+$Percept   & $A+Q+K+P$   & Observations also supplied \\
\midrule
\ding{175} No audio     & $Q+K$       & Audio-necessity control \\
\bottomrule
\end{tabular}
\end{table}

Condition~\ding{175} probes whether the question and rule support answering without audio.
Its direct audio ablation is relative to \ding{173}; comparing \ding{174} with \ding{175}
changes both audio and percept text. A single correct no-audio answer may be a guess and
does not establish leakage. We therefore report aggregate control accuracy against the
chance rate implied by the choices. The present results do not certify individual items as
free of text-only shortcuts (Section~\ref{sec:validation}).

\subsection{Operational diagnostic assignment}\label{sec:diagnosis}
Let $z_j$ indicate whether the response under diagnostic condition $j$ is correct. We assign
each model--item pair to the first successful condition in the order \ding{172}, \ding{173},
\ding{174}. The correctness patterns $(z_1,z_2,z_3)$ define four disjoint groups:
\begin{itemize}
  \item $1**$: \textbf{capable}, correct without supplied information;
  \item $01*$: \textbf{missing knowledge}, first correct with the rule supplied;
  \item $001$: \textbf{failed perception}, first correct with observations also supplied;
  \item $000$: \textbf{failed application}, incorrect under all three conditions.
\end{itemize}
Here $*$ denotes either outcome. Fig.~\ref{fig:pipeline}(b) visualizes this assignment.
The labels summarize responses to interventions; they do not establish a unique internal
cause or identify a particular model component. In particular, an item answered correctly
under \ding{172} remains in the capable group even if a later condition is incorrect.
Non-monotonic responses must therefore be examined alongside the four group sizes.

The intended interpretation requires a correct, usable rule and sufficient, accurate percept
text. Supplying a rule may also change prompting or attention, and a correct multiple-choice
answer need not reflect a successful reasoning chain. Likewise, an incomplete $P$ can leave
perception necessary under \ding{174}, whereas a $P$ that states the answer can bypass
application. Earlier builds contained event-rate percepts without duration and category percepts
that named the answer. These examples motivate checking the sufficiency of $P$ and whether it
preserves an application step. The current results should be read with the validation coverage
reported in Section~\ref{sec:perceptcheck}.

\subsection{Item schema}
Each item is stored as a record carrying its identifier and family; the source, license, and a
contamination flag; the \texttt{question}, \texttt{choices}, and gold answer; the
\texttt{knowledge} string; and \texttt{percept\_name} and \texttt{percept\_text}, the latter being
the sentence inserted at condition \ding{174}. The four materialized prompts are stored alongside,
so a third party can reproduce the exact input the model saw without rerunning the pipeline.
The local build associates one audio clip with each item.

The proposed release separates item records and analysis code from source audio. Audio packaging
will follow the terms of each source, including for transformed recordings. Where redistribution
is not permitted, the intended package provides source identifiers, time spans and transformation
instructions for local reconstruction. Source-specific permissions and the final licences for
release materials remain to be confirmed; this manuscript does not constitute a released dataset.

\subsection{Contamination}\label{sec:contamination}
Items carry contamination-risk flags. Newly synthesized audio reduces the risk of exposure to
the exact clip, but does not eliminate shortcuts in question templates or answer choices.
Research corpora without published question--answer pairs are marked as lower risk; this is
a provenance-based flag, not a measured guarantee.

The current build retains one MMAR well-depth item whose question, answer and explanation are
public. It is included in the reported totals and marked as high risk. We retain it as a case
study, not as a validated positive control. A single item cannot establish the power of the
no-audio condition to detect contamination. That requires a separate contamination set and
comparisons with unflagged items.

\section{Data and construction}\label{sec:sources}
This section describes where AuDiR's audio and its answers come from. One requirement governs
the whole procedure: an item's answer must be checkable against a measurement that the audio
did not produce. That requirement, rather than audio quality or item count, is what limits the
size of the benchmark, and it holds per target quantity rather than per corpus, so the same
recordings are admitted for some questions and rejected for others. We report the sources in
the current build, the composition they yield, the construction procedure that turns an admitted
pairing into items, the pairings that procedure rejected, and the families being recorded to
replace them.

\subsection{Sources in use}
Table~\ref{tab:sources} lists the \NSources{} sources behind the current build. They fall into
four kinds according to where the answer comes from: an instrument recorded beside the microphone,
a physical measurement of the object, an annotation produced outside any acoustic estimator, or a
quantity set by the experimenter before the audio existed. What follows states, for each source,
what the recordings are, how its answer reference was obtained, and what we did and did not build
from it. None of these corpora was assembled for reasoning evaluation; each is admitted for the
specific quantities whose provenance survives inspection.

\textbf{VS13}~\cite{vs13} records single vehicles driving past a camera and microphone at the
roadside, thirteen passenger cars chosen to differ in manufacturer, year, engine and
transmission, over four hundred passes between 30 and 105\,km/h. The speed of each pass is the
speed the driver held on the car's cruise control, which the corpus reports is stable for at
least three seconds either side of the pass-by; it is logged per file, together with the instant
the car leaves the camera frame, which approximates the closest point of approach. That speed is
set before the recording and is not derived from the sound, which is what admits the source, but
it is the vehicle's own reading of itself rather than a roadside measurement, and a speedometer
is permitted to read high. We build the two Doppler families from it and use no other quantity.

\textbf{Sound of Water}~\cite{soundofwater} records water being poured into fifty-one vessels,
each measured with a ruler: top and bottom internal diameter, net internal height, and for some
the wall thickness. Each of its recordings names the vessel it used and carries flags for a
constant pour and a clean recording, both of which we require. The geometry is the answer
reference for both families we build, container height and the taller/fuller comparison, and it
is a measurement of the object rather than of the sound. The corpus ships no pitch annotation, so
the percept is measured by us and checked against the geometry rather than the other way round.
Two thirds of the vessels are visibly tapered, which is why volume is computed as a truncated cone
and why the comparison family states the radius in the question.

\textbf{GuitarSet}~\cite{guitarset} records solo and accompaniment guitar takes with a hexaphonic
pickup, one channel per string, alongside the ordinary mixed recording. The note, chord, beat and
key annotations are derived from the per-string channels, so they do not come from the signal a
model is given. We build three families on the mixed recording: the interval between two notes,
the quality of a four-note chord, and the major key of a passage. The same recordings also supply
the unmodified notes that the detuning family transforms.

\textbf{AudioTime}~\cite{audiotime} is constructed audio in which the onset and offset of every
sound event are fixed when the clip is assembled rather than annotated afterwards, which makes the
event timestamps exact by construction. We use the event counts and clip durations for the two
event-rate families. We do not use the timestamps themselves as an answer, because a question
about when something happened is answered by reading the timestamp rather than by applying a rule.

\textbf{MAFAULDA}~\cite{mafaulda} records a rotating machine under induced fault conditions with a
tachometer channel recorded simultaneously with the microphones. The tachometer is a sensor
independent of the microphone, which makes the rotation rate the strongest ground truth in the
build. We use it for absolute rotation speed and the speed ratio between two machines. We do not
build the fault-diagnosis question the corpus was made for: the bearing fault frequencies sit
within a fraction of a hertz of the shaft harmonics at these speeds, which fails C5.

\textbf{Hillenbrand vowels}~\cite{hillenbrand} records isolated American English vowels read by
men, women, boys and girls at 16\,kHz, with formant tracks supplied every 10\,ms. The vowel label
is what the speaker was asked to produce, not what an analyzer decided afterwards, so it is
independent of any acoustic measurement. We build vowel identity from the first two formants and
a formant chart stated per speaker group.

\textbf{CirCor DigiScope}~\cite{circor} records heart sounds from nearly a thousand paediatric
patients at up to four auscultation sites, with each recording segmented by clinicians into the
first heart sound, systole, the second heart sound and diastole. The segmentation is expert
annotation rather than an algorithm's output. We build heart rate from the instants of the heart
sounds, which requires the rule that a beat produces two sounds and that the rate follows the gap
between successive first sounds, not between successive sounds.

\textbf{Controlled transformation of a real recording} supplies the families whose target
quantity does not exist in any corpus we could find. A recorded engine is flown past a virtual
microphone at a chosen speed and distance; a recorded guitar note is detuned by a chosen number of
cents against an unmodified copy of itself; a recorded pour or pass-by is band-limited to a chosen
sampling rate and converted back up. The transformation parameter is exact because it was fixed
before the audio was rendered, and the sound remains a real engine, guitar or pour rather than a
synthesized tone. \NSynth{} of \NItems{} items are built this way. The limitation is stated in
Section~\ref{sec:limitations}: a rule that the renderer applied is a rule the item cannot test for
truth, only for application.

\textbf{MMAR}~\cite{mmar} contributes one item, retained deliberately. Its question, answer and
chain of thought are public, so it is flagged as contaminated, counted in every total, and read as
a case study rather than as a positive control on the leak test.

Licences differ, and one source does not permit redistribution of audio, so the release is partly
by fetch script rather than a single archive.

\begin{table*}[t]
\centering
\caption{Sources currently in use. A source is admitted per pairing, not as a whole, so the same recordings support some target quantities and not others. Answer references are listed by their documented provenance; shipped annotations still require an independence audit. The MMAR item is retained as a flagged case study.}
\label{tab:sources}
\scriptsize
\begin{tabular}{@{}>{\raggedright\arraybackslash}p{2.70cm}>{\raggedright\arraybackslash}p{4.40cm}>{\raggedright\arraybackslash}p{4.40cm}>{\raggedright\arraybackslash}p{3.05cm}rr@{}}
\toprule
Source & Audio & Answer reference & License & Fam. & Items \\
\midrule
VS13 & single vehicles passing a roadside microphone at constant speed & speed held on the vehicle's cruise control, logged per pass & IEEE, research use, no redistribution & 2 & 24 \\
Sound of Water & water poured into vessels of known geometry & ruler-measured net height and container geometry & MIT & 2 & 20 \\
GuitarSet & solo acoustic guitar, hexaphonic pickup & shipped chord and note transcription & CC BY 4.0 & 3 & 36 \\
AudioTime & environmental scenes with several distinct sound events & shipped onset and offset timestamps & open & 2 & 24 \\
MAFAULDA & rotating machinery under controlled fault conditions & tachometer channel recorded alongside the audio & open (UFRJ) & 2 & 24 \\
MMAR & one retained clip, a stone dropped into a well & answer shipped with the benchmark & CC BY-NC 4.0 & 1 & 1 \\
Hillenbrand vowels & isolated American English vowels read by men, women and children & corpus vowel label with measured formants & research use & 1 & 12 \\
CirCor DigiScope & paediatric heart sounds recorded at four auscultation sites & expert S1 and S2 segmentation & open (PhysioNet) & 1 & 12 \\
Constructed from real recordings & real engine, guitar and pouring recordings put through a controlled transformation & the transformation parameter, fixed before rendering & follows the source recording & 4 & 48 \\
\midrule
\multicolumn{4}{@{}l}{\textbf{Total}} & \textbf{18} & \textbf{201} \\
\bottomrule
\end{tabular}
\end{table*}

\subsection{Families and perceptual axes}\label{sec:structure}
Table~\ref{tab:composition} lists the \NFamilies{} families these sources yield, including the
flagged contamination case of Section~\ref{sec:contamination}. Together they hold
\AudioMinutes{} minutes of audio (mean \AudioMean{}\,s, median \AudioMedian{}\,s, range
\AudioMin{}--\AudioMax{}\,s). All \NFourWay{} items have four choices, giving a uniform-guessing accuracy of \Chance{}\%. The question forms comprise
\NAbsolute{} absolute estimates, \NCategorical{} classifications and \NComparative{} comparisons
or ratios. Counts and durations refer to the current item set.

\begin{table*}[t]
\centering
\caption{Current AuDiR composition. Families pair acoustic observations with a target quantity and a stated rule. The reference column records the documented answer source; independence of shipped annotations requires provenance verification. Counts include the flagged MMAR case study and remain subject to revision.}
\label{tab:composition}
\scriptsize
\begin{tabular}{@{}>{\raggedright\arraybackslash}p{2.45cm}>{\raggedright\arraybackslash}p{1.95cm}>{\raggedright\arraybackslash}p{4.30cm}>{\raggedright\arraybackslash}p{3.55cm}>{\raggedright\arraybackslash}p{3.30cm}r@{}}
\toprule
Family & Source & Percept (heard) & Rule (told) & Answer reference & $n$ \\
\midrule
Doppler speed & VS13 & approach and recede engine-tone frequencies & $f = f_0 c/(c \mp v)$ & cruise-control speed & 12 \\
Doppler speed ratio & VS13 & both cars' tone shifts & ratio of two Doppler solutions & cruise-control speed & 12 \\
Container height & Sound of Water & empty-container resonant pitch & $f = c/\bigl(4(H+\beta R)\bigr)$ & ruler-measured net height & 10 \\
Taller vessel, fuller vessel & Sound of Water & both containers' empty-start pitches & $f \to H$, then $V$ from $H$ and the given $R$ & ruler height and diameters & 10 \\
Event rate & AudioTime & number of distinct sound events & $\mathrm{rate} = 60N/T$ & shipped event timestamps & 12 \\
Chord quality & GuitarSet & the chord's interval pattern & stacked thirds $\to$ quality & GuitarSet transcription & 12 \\
Melodic interval & GuitarSet & the two pitches heard & semitone distance $\to$ interval name & GuitarSet note transcription & 12 \\
Shaft rotation speed & MAFAULDA & shaft-rotation tone & $\mathrm{RPM} = 60 f$ & tachometer channel & 12 \\
Well depth & MMAR & release-to-splash delay & $H = \tfrac{1}{2}gT_1^2$ plus return at $c$ & dataset-provided answer & 1 \\
Rotation speed ratio & MAFAULDA & two shaft-rotation tones & ratio of rotation rates & tachometer & 12 \\
Event rate ratio & AudioTime & event counts and clip lengths in two recordings & ratio of counts per minute & shipped timestamps & 12 \\
Pass-by distance & generated & approach and recede tones and the steepest slope & Doppler pass-by geometry & construction parameter & 12 \\
Motion direction & generated & engine tone and level at three instants & Doppler shift versus inverse-square level & construction parameter & 12 \\
Original sampling rate & generated & the highest frequency still present & Nyquist limit & construction parameter & 12 \\
Vowel identity & Hillenbrand & the first two formants & formant chart for the speaker group & corpus label & 12 \\
Major key & GuitarSet & the pitch classes used & major-scale membership & shipped key annotation & 12 \\
Heart rate & CirCor & the instants of the heart sounds & two sounds per beat, rate from the S1-to-S1 gap & expert segmentation & 12 \\
String detuning & generated & the rate at which loudness pulses & beat rate equals the frequency difference & construction parameter & 12 \\
\midrule
\multicolumn{5}{l}{\textbf{Total}} & \textbf{201} \\
\bottomrule
\end{tabular}
\end{table*}

A \emph{family} is an item template: one source, one percept, one rule. A \emph{perceptual axis}
groups the observations to be recovered, such as a pitch pattern, an event time or a decay tail.
An axis can group multiple families, providing a way to organize response profiles by
perceptual demand. In the current build, most axes occur in only one source, so axis effects
cannot generally be separated from source effects. Table~\ref{tab:facets} reports this
organization, not evidence of cross-source generalization.

\begin{table}[t]
\centering
\caption{Distribution over perceptual axes, question forms, and rule domains. Axes organize items by perceptual demand; most currently occur in only one source.}
\label{tab:facets}
\small
\begin{tabular}{lr}
\toprule
Category & $n$ \\
\midrule
\multicolumn{2}{l}{\textit{Percept type}} \\
\quad frequency shift & 24 \\
\quad event count & 24 \\
\quad rotation tone & 24 \\
\quad pitch pattern & 24 \\
\quad resonant pitch & 20 \\
\quad frequency trajectory & 12 \\
\quad level trend & 12 \\
\quad bandwidth & 12 \\
\quad formants & 12 \\
\quad pitch classes used & 12 \\
\quad beat rate & 12 \\
\quad cycle timing & 12 \\
\quad delay / timing & 1 \\
\addlinespace
\multicolumn{2}{l}{\textit{Question form}} \\
\quad absolute & 95 \\
\quad categorical & 60 \\
\quad comparative & 46 \\
\addlinespace
\multicolumn{2}{l}{\textit{Rule domain}} \\
\quad physics & 105 \\
\quad music & 36 \\
\quad counting & 24 \\
\quad signal & 12 \\
\quad phonetics & 12 \\
\quad clinical & 12 \\
\bottomrule
\end{tabular}
\end{table}

\subsection{Construction pipeline}\label{sec:pipeline}
\begin{figure*}[t]
\centering
\includegraphics[width=\textwidth]{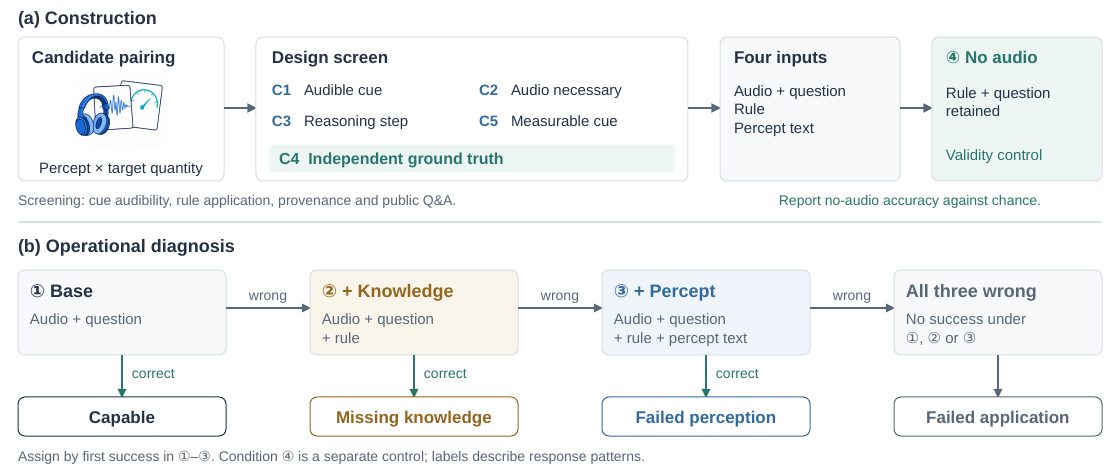}
\caption{AuDiR construction and diagnosis. (a)~Candidate percept--target pairings are screened
and materialized into four inputs. The audio-removed condition~\ding{175} probes text-only
answerability; current control results are reported in aggregate. Rejection reasons are detailed
in Table~\ref{tab:rejected}. (b)~The first correct response among conditions
\ding{172}--\ding{174} assigns the operational diagnosis; failure under all three is
assigned to application. Condition~\ding{175} is a separate validity control,
not a fourth step in this assignment sequence.}
\label{fig:pipeline}
\end{figure*}

Fig.~\ref{fig:pipeline} summarizes admission and the subsequent diagnostic assignment.
The pipeline has seven stages. Its binding constraint is not audio quality or item count but
\emph{ground-truth provenance}, and most candidate work is eliminated at stage two. One property
guides construction: the percept text and gold answer should follow separate derivation paths.
What can be heard determines condition~\ding{174}'s text; an independent reference determines
the answer. This avoids treating agreement with an audio estimator as independent correctness.
The provenance of shipped annotations requires further documentation to establish this
separation for every current family.

\subsubsection{Candidate sourcing as percept--target pairing}
We select candidate percept--target pairs from available recordings. A candidate is one (percept, target
quantity) pairing: something a human can hear, and something reachable from it by a stated rule.
One recording source can yield several pairings, and they are admitted or rejected
independently. Sound of Water, for instance, supports container height but not fill rate, and
Table~\ref{tab:sources} and Table~\ref{tab:rejected} list the same corpus on both sides.

\subsubsection{Admission test}
Candidate pairings are screened against the following design criteria:

\begin{itemize}
  \item[\textbf{C1}] \textbf{Percept present}. the cue the rule consumes is actually audible to
        a human, at the recording's bandwidth and noise floor.
  \item[\textbf{C2}] \textbf{Audio-necessary}. the answer cannot be reached from the question
        text, the file name, or dataset priors.
  \item[\textbf{C3}] \textbf{Real reasoning step}. the rule must be applied. A pairing where
        the percept \emph{is} the answer tests recognition, not reasoning.
  \item[\textbf{C4}] \textbf{Independent ground truth}. the answer is checked against a
        measurement that does not derive from the signal being analyzed.
  \item[\textbf{C5}] \textbf{Measurable}. the percept can be recovered from the audio reliably
        enough that a correct chain reaches the correct answer.
\end{itemize}

\textbf{C4 is the criterion that rejects most candidates}, and it deserves emphasis because it is
easy to violate accidentally. A ground truth estimated from the same signal the model is asked to
analyze is \emph{circular}: the item then certifies only that our estimator and the model agree,
and a model that hears perfectly but disagrees with our estimator is scored wrong. An item can satisfy the other design criteria while still relying on an inaccurate reference.
Reference provenance therefore needs a separate check.

Table~\ref{tab:rejected} records rejected candidate pairings and the criteria they failed.

\subsubsection{Percept extraction and ground-truth binding}
For an admitted pairing we extract the percept value from the audio and bind the answer to the
independent ground truth. These are two separate operations on two separate sources.
The percept is what a careful listener could obtain, and it is used only to write condition
\ding{174}'s text. The answer comes from the instrument, annotation, or construction listed in
Table~\ref{tab:composition}. The ground truths in AuDiR are of four
kinds. An \emph{instrument} independent of the microphone (a tachometer channel recorded beside
it), a physical \emph{measurement} of the object (container geometry by ruler), a
shipped \emph{annotation} produced outside our pipeline (event timestamps, note
transcription), and \emph{control}, where the quantity is set before the audio exists, either by
the experimenter during recording (a driver holding cruise control at a stated speed) or by
rendering (a pass-by synthesized at a chosen speed and distance). A control is only as good as
the setting is faithful: a cruise-control speed is the vehicle's own reading of itself, and a
speedometer is permitted to read high. Being external to our pipeline does not establish that an annotation is independent
of the analyzed signal; that provenance must be audited. The retained MMAR case has only a
dataset-provided answer documented here and should not be read as an instrument-validated item.

\subsubsection{Percept extraction and verification}\label{sec:perceptcheck}
The percept sentence is produced by a measurement routine written for the family, not by a person
listening and describing. Pitches and pitch shifts come from a fundamental-frequency tracker run
over the clip and read at the instants the rule needs. Event counts and instants come from an
onset detector, bounded by the source's own timestamps where it ships them. Levels come from
short-time energy in decibels relative to full scale. Bandwidth comes from the highest frequency
carrying appreciable energy in the long-term spectrum. Formants come from linear prediction.
The routine writes one sentence per item from a fixed template, so the wording varies only where
the numbers do.

Errors in percept text can affect the interpretation of condition~\ding{174}. We use three
checks: applying the stated rule to the percept, re-measuring the audio using another method,
and comparison with a standard tool where available. The stored check records cover
\NChkRuleCovered{} of \NItems{} current items; the \NChkRuleMissing{} heart-rate items have no
records in these three checks.

Rule application agrees with gold on \NChkRuleAgree{} items and disagrees on
\NChkRuleDisagree{}, with \NChkRuleNA{} marked not applicable. Re-measurement agrees on
\NChkRemeasAgree{}, disagrees on \NChkRemeasDisagree{}, and is not applicable on
\NChkRemeasNA{}. The tool comparison agrees on \NChkToolAgree{}, disagrees on
\NChkToolDisagree{}, and is not applicable on \NChkToolNA{}. Non-applicability is distinct from
a passed check. The disagreements and missing coverage require further review; these results
do not establish that every current percept has been validated.

A separate check compares stored gold options with the bins implied by the recorded reference
values. It finds \NGoldAgree{} agreements and \NGoldDisagree{} disagreements, with
\NGoldNA{} items outside this numeric check. This verifies consistency between reference values
and answer bins; it does not by itself establish that the reference was measured independently
of the audio.

A worked case shows what the checks are for. One family asks which of two vessels is taller and
which holds more, from the pitch each makes while filling. In an earlier build both halves of the
answer were derived from that pitch, which made the gold answer a measurement of the audio rather
than an independent one, and on one pair it was simply wrong: two vessels whose ruler heights
differ by one centimetre, against a pitch estimate carrying several percent of error, came out in
the wrong order. The repair was to take the answer from the ruler alone, to compute volume as a
truncated cone rather than a cylinder because only ten of the corpus's fifty-one vessels are
straight-sided to within five percent, and to drop the pairs whose heights are closer together
than the pitch can resolve. Nothing about the audio changed; what changed is that the answer no
longer comes from the same signal the model is asked to listen to.

These checks address different questions. Rule application checks whether the supplied
information leads to the reference answer. Re-measurement and tool comparisons assess whether
that information describes the recording. Agreement among them is useful evidence, but it does
not replace a documented provenance audit or a human listening study.

\subsubsection{Rule authoring}
The knowledge string is written once per family and reused across its items. It is checked for
item-specific observations and answers. It states the
governing relation and the constants it needs, and stops there. Where a quantity is required that
is neither hearable nor general, such as a container's radius, it is supplied in the
\emph{question}, as a given, in every condition alike. This keeps the tell/hear split clean: the
question may carry item-specific constants, the knowledge may not, and neither may carry the
percept.

\subsubsection{Option construction}
Items are multiple choice so that a wrong answer is unambiguous and chance is known.
All \NFourWay{} current items have four choices, giving a uniform-guessing chance rate of \Chance{}\%. Options are built in one of three forms
according to the question: contiguous \emph{bins} over the continuous target for absolute
estimates, the natural \emph{class labels} for classification, and \emph{ratio bands} for
comparisons. Bin
edges are cut per family, from that family's own ground-truth values, by a rule with three parts.
The three interior edges are placed at midpoints between adjacent ground-truth values; among the
placements that keep the four bins near-equal in count, the one chosen is the one that maximizes
the smallest distance from any ground-truth value to any edge. The two outer bins are closed and
given the width of their neighbour, widened if needed to cover the data. This increases the margin around the reference values and reduces sensitivity to small
numerical errors.

One hard constraint governs the cut: the value obtained by applying the family's rule to the
percept sentence alone must land in the same bin as the ground truth. If it does not, either the
bins or the item are wrong, and condition~\ding{174} would be unanswerable even for a model that
did everything asked of it. That constraint is checked for every numeric item
(Section~\ref{sec:perceptcheck}). The gold option's position is then balanced across the build so
that answer position carries no signal.

\subsubsection{Materialization and leak validation}
Each admitted item is finally rendered into the four prompts of Table~\ref{tab:conditions} from
the same fields. Condition~\ding{175} probes whether the rule and question admit a text-only
shortcut. The current implementation reports aggregate no-audio accuracy and does not
perform a statistically validated per-item rejection test. Such a gate would require an
explicit evaluation protocol, decision threshold and screening record; a single correct
multiple-choice response is not sufficient evidence for discarding an item. Consequently,
the current build should not be described as certified free of text-only leakage.


\subsection{Pairings we rejected}
\NRejected{} pairings were considered and not admitted (Table~\ref{tab:rejected}). The pattern is
worth stating because it is the main practical finding of the construction effort: audio that is
easy to hear and governed by a clean rule is common, and audio whose answer can be checked against
something other than itself is rare.

C4 is the single most common cause of rejection, and it fails in two distinct ways. Some candidates fail for
\emph{circularity}: the water level in a filling vessel follows from the resonance the model is
being asked to interpret, so a ground truth derived that way would certify only that our estimator
and the model agree. Others fail for \emph{absence}: the delay between a gunshot's shockwave and
its muzzle blast is a clean percept with a clean rule, and no gunshot corpus we found records the
shooter's range. The two are worth separating because only the second can be fixed by finding a
better corpus.

The remaining rejections are cheaper to explain. A single-room corpus gives the target quantity one
value across every item, so the question has a constant answer. Firearm-model identification is
recognition rather than reasoning. Echo delays of two to four milliseconds are below what a
listener can resolve, so the percept is not there to be heard whatever the ground truth says.

\begin{table}[t]
\centering
\caption{Pairings considered and not admitted. C4, an independent ground truth, is the single most common cause and the only one that cannot be repaired by listening harder or recording better. A rejection removes one target quantity, not the source: Sound of Water appears here three times and in Table~\ref{tab:sources} as an admitted source.}
\label{tab:rejected}
\scriptsize
\begin{tabular}{@{}>{\raggedright\arraybackslash}p{1.50cm}>{\raggedright\arraybackslash}p{2.15cm}>{\raggedright\arraybackslash}p{1.05cm}>{\raggedright\arraybackslash}p{3.05cm}@{}}
\toprule
Source & Target quantity & Fails & Reason \\
\midrule
Sound of Water & water level or fill fraction & C4 & would be read back from the same pitch trajectory the model is given \\
Sound of Water & fill rate in mL/s & C4 & no flow meter accompanies the recordings \\
Sound of Water & time remaining until full & C4 & circular, and no vessel is filled to the top \\
dEchorate & source distance from echo delay & C1, C5 & first-reflection delays are 2--4\,ms, below what is audible \\
dEchorate & direction of arrival & C3 & needs multichannel input and resolves in one step \\
dEchorate & room size & C3 & the corpus has one room, so the answer is constant \\
Gunshot corpora & firearm model & C3 & recognition in one step, no rule to apply \\
Gunshot corpora & shooter distance & C4 & no measured range is recorded \\
Gunshot corpora & rate of fire & C3, C4 & one step, and read off the same timestamps \\
MMAR & dive height, water-level ordering, tempo & contam. & question, answer and chain of thought are public \\
RealImpact & struck-object material & C5 & decay reaches the noise floor within ${\sim}0.2$\,s; parked pending a replacement corpus \\
\bottomrule
\end{tabular}
\end{table}

\subsection{Sources under construction}
The scarcity described above is the reason the benchmark is being extended by recording rather
than by searching. Table~\ref{tab:planned} lists \NPlanned{} families in progress. Each is set up
so that the target quantity is measured by an instrument that never touches the audio, or is fixed
when the stimulus is generated: a ruler for a pendulum's length, hanging calibrated weights for a
string's tension, a laser tachometer for a rotating wheel. Recording our own audio is slower than
reusing a corpus and it is the only way we have found to satisfy C4 at scale.

\begin{table}[t]
\centering
\caption{Families under construction. Each is recorded or synthesized so that the target quantity is fixed or measured by an instrument that never touches the audio, which is what C4 requires and what most existing corpora cannot supply.}
\label{tab:planned}
\scriptsize
\begin{tabular}{@{}>{\raggedright\arraybackslash}p{2.05cm}>{\raggedright\arraybackslash}p{2.85cm}>{\raggedright\arraybackslash}p{3.00cm}@{}}
\toprule
Family & Target quantity & Independent ground truth \\
\midrule
Bouncing ball & coefficient of restitution, drop height & high-speed video and ruler \\
Free fall into a shaft & drop height & laser rangefinder \\
Pendulum & string length & ruler \\
Rotating wheel & surface speed, revolutions per second & laser tachometer and ruler \\
Struck rod & rod length, material & ruler; alloy known at capture \\
Plucked string & tension & hanging calibrated weights \\
Helmholtz resonator & cavity volume & measuring cup \\
Resonance tube & tube length & ruler \\
Synthesized beats & detuning in cents, tension change & set by construction \\
\bottomrule
\end{tabular}
\end{table}

\section{Experimental setup}\label{sec:setup}
\subsection{Model}
We report a preliminary evaluation of \MAName{}, accessed through an audio-capable API, over
all \NItems{} items and four conditions. This is the only model with complete coverage in the
reported results. Evaluations of additional models are in progress and are not included in the
comparisons. The protocol requires joint audio and text input and a textual answer, without
access to model weights or intermediate representations.

\subsection{Prompting}\label{sec:prompting}
Every condition uses one user turn and no system message. The turn is assembled from at most four
parts in a fixed order: the rule, the percept sentence, the question, and the options, each
separated by a blank line. Condition~\ding{172} supplies the question and options only.
Condition~\ding{173} prepends the rule. Condition~\ding{174} prepends the rule and then the percept
sentence, introduced by the words \texttt{Given (measured from the sound):}, so the model can tell
the supplied observation from the rule. Condition~\ding{175} is byte-identical to \ding{173} with
the audio file removed from the request. Options are lettered A to D and presented in a fixed order
that does not depend on the condition, so the four passes over an item differ only in what is
prepended and whether audio is attached.

Each turn asks the model to work through the problem and end with a final line reading
\texttt{Answer: X}. The same instruction is used in all four conditions. No few-shot examples
are provided, and the template is not tuned separately for each family.

\subsection{Decoding and scoring}\label{sec:scoring}
We use temperature $0$ and one response per item and condition. These are single-run estimates;
the stability of the response labels across repeated API calls has not been measured.
The evaluation records truncated and unparseable responses separately. The complete run reported
here contains neither, so every condition is scored over all \NItems{} items.

A completed response is scored using the letter on its final \texttt{Answer:} line. If that line
is absent, the parser uses the last standalone option letter; if no letter is found, the response
is scored incorrect. No partial credit is given, and the reasoning text is not scored. We report
accuracy with 95\% Wilson intervals. These intervals summarize item-level outcomes within this
build; they do not account for dependence among items from the same family.

\subsection{Baseline validation}\label{sec:validation}
\begin{table}[t]
\centering
\caption{Baseline check on 201 items. Brackets are 95\% Wilson intervals; chance is 25.0\%. The reported no-audio interval contains chance. This aggregate check does not establish the absence of text-only shortcuts in individual items.}
\label{tab:validation}
\begin{tabular}{lc}
\toprule
Condition & Gemini\nobreakdash-3.8\nobreakdash-flash \\
\midrule
\ding{172} & 29.3 \tiny{[23, 36]} \\
\ding{173} & 31.3 \tiny{[25, 38]} \\
\ding{174} & 98.5 \tiny{[96, 99]} \\
\midrule
\ding{175} & 23.9 \tiny{[19, 30]} \\
\midrule
\emph{Chance} & \multicolumn{1}{c}{25.0} \\
\bottomrule
\end{tabular}
\end{table}

\textbf{Aggregate no-audio accuracy is consistent with chance.} With the audio removed but the rule and question intact,
\MAName{} scores \MANoaudio{}\%, with a 95\% Wilson interval containing the
\Chance{}\% chance rate (Table~\ref{tab:validation}). This check does not establish the
absence of shortcuts in individual items, nor does it rule out text-only performance by
other models. The current results are aggregate baseline evidence, not a per-item leakage certificate.

\textbf{Percept text produces the larger aggregate gain.} Supplying the rule (\ding{173})
changes accuracy only modestly, while supplying the percept (\ding{174}) raises it to \MAPercept{}\%. Since \ding{174} differs from \ding{173} only in that the heard quantity is
written out, this contrast measures the benefit of the supplied percept representation.
Interpreting it as a pure perception deficit additionally requires that the text be complete and
preserve the intended application step. Aggregate changes also conceal non-monotonic item-level
responses, reported below.

\subsection{Knowledge-recall probe}\label{sec:probe}
A small aggregate gain from supplying the rule could reflect prior knowledge, limited use of
the supplied text, or offsetting improvements and regressions. To investigate prior knowledge,
we ask one question per family requesting the governing relation and its symbols, without audio,
answer choices, or the reference rule. A separate call to the same model, \MAName{}, grades the
response against the family rule as equivalent, partial, or different. The grading prompt omits
the respondent's identity, but using the same model for answering and grading remains a
limitation; these labels have not been independently validated by human annotators.

\begin{table}[t]
\centering
\caption{Family-level rule recall for Gemini\nobreakdash-3.8\nobreakdash-flash, graded by a separate call to the same model. Columns report accuracy (\%) under each condition; $\Delta$ is the change from base to rule-supplied accuracy in percentage points. Groups describe probe responses, not verified internal knowledge states.}
\label{tab:probe}
\small
\setlength{\tabcolsep}{3pt}
\begin{tabular}{@{}lrrrrrr@{}}
\toprule
Probe outcome & Fam. & $n$ & \ding{172} & \ding{173} & $\Delta$ & \ding{174} \\
\midrule
states the same rule & 10 & 107 & 25.2 & 25.2 & +0.0 & 97.2 \\
states it incompletely & 5 & 58 & 20.7 & 32.8 & +12.1 & 100.0 \\
states a different rule & 3 & 36 & 55.6 & 47.2 & -8.3 & 100.0 \\
\bottomrule
\end{tabular}
\end{table}

Table~\ref{tab:probe} groups item accuracy by the family-level probe outcome. The
\NProbeEquivalent{} families graded equivalent have a net change of \ProbeGainEquivalent{}
percentage points after rule provision; the \NProbePartial{} families graded partial improve
by \ProbeGainPartial{} points. This association is consistent with greater benefit when the
recalled rule is incomplete. It does not prove that the supplied rule was used in a particular
response, and zero net change can conceal opposing item-level transitions.

The \NProbeWrong{} families graded different show a net change of \ProbeGainWrong{} points.
A response in this group may state an alternative relation or an incorrect one; the grading
scheme does not establish that it is a valid solution route. The probe is therefore supplementary
evidence about rule recall, not an independent identification of each item's failure cause.

\section{Analysis}\label{sec:analysis}

\subsection{Failure profiles}
\begin{table}[t]
\centering
\caption{Operational attribution over all 201 current items, including the flagged case. An item is assigned to the first condition at which the model answers it correctly: \ding{172} nothing supplied, \ding{173} fixed by supplying the rule, \ding{174} fixed by supplying the percept. An item incorrect under all three diagnostic conditions is assigned to application. Unaided accuracy is 29.3\%; the labels summarize intervention responses.}
\label{tab:bottleneck}
\small
\begin{tabular}{@{}lrr@{}}
\toprule
& \multicolumn{2}{c@{}}{Gemini\nobreakdash-3.8\nobreakdash-flash} \\
\cmidrule(l){2-3}
Diagnosis & $n$ & \% \\
\midrule
\ding{172}~Capable & 59 & 29.3 \\
\ding{173}~Missing knowledge & 20 & 10.0 \\
\ding{174}~Failed perception & 122 & 60.7 \\
\hphantom{\ding{172}}~Failed application & 0 & 0.0 \\
\midrule
Total & 201 & 100 \\
\bottomrule
\end{tabular}
\end{table}

Table~\ref{tab:bottleneck} assigns every item to the first condition at which the model answers it
correctly, and assigns items incorrect under all three diagnostic conditions to application.
These are operational labels as defined in Section~\ref{sec:diagnosis}.

Of the \NItems{} items, \BtNMACapable{} are correct without supplied information,
\BtNMAKnowledge{} first become correct with the rule, and \BtNMAPerception{} first become
correct with percept text also supplied. No item is incorrect in all three diagnostic conditions.
This last result does not mean that application never fails: condition~\ding{174} still has
\NMAFullWrong{} incorrect answers, each assigned to an earlier group because it was correct
under an earlier condition.

Responses are not monotonic. There are \NMABaseRuleReversal{} items that are correct under
\ding{172} but incorrect under \ding{173}; \NMAAnyReversal{} show a correct-to-incorrect
reversal at either adjacent transition. These patterns limit interpretations of the labels as
stable internal failure mechanisms. In particular, supplying percept text is not equivalent to
improving an audio encoder, so the size of the perception group does not estimate the gain from
an encoder replacement.

Sensitivity analyses excluding the flagged contamination case and transformed recordings
remain to be completed.

\subsection{What the failures depend on}\label{sec:axes}
\begin{table}[t]
\centering
\caption{First-success counts for Gemini\nobreakdash-3.8\nobreakdash-flash by perceptual axis. The last column is the percentage first solved with percept text. The single flagged well-depth item is excluded from this breakdown; it remains in the overall results.}
\label{tab:axes}
\small
\begin{tabular}{@{}lrrrrr@{}}
\toprule
Percept axis & $n$ & \ding{172} & \ding{173} & \ding{174} & \ding{174}\% \\
\midrule
resonant pitch & 20 & 5 & 0 & 15 & 75 \\
cycle timing & 12 & 3 & 0 & 9 & 75 \\
rotation tone & 24 & 4 & 3 & 17 & 71 \\
frequency trajectory & 12 & 1 & 3 & 8 & 67 \\
level trend & 12 & 3 & 1 & 8 & 67 \\
beat rate & 12 & 3 & 1 & 8 & 67 \\
event count & 24 & 7 & 2 & 15 & 62 \\
frequency shift & 24 & 5 & 5 & 14 & 58 \\
pitch pattern & 24 & 7 & 3 & 14 & 58 \\
bandwidth & 12 & 4 & 2 & 6 & 50 \\
pitch classes used & 12 & 7 & 0 & 5 & 42 \\
formants & 12 & 10 & 0 & 2 & 17 \\
\bottomrule
\end{tabular}
\end{table}

Table~\ref{tab:axes} groups the first-success labels by perceptual axis. Resonant pitch and
cycle timing have the largest shares first solved with percept text, while vowel identity has
the largest share solved without supplied information.

Some categorical questions may permit direct recognition. For example, a model can name a
vowel without explicitly reporting formants, or identify a musical key without enumerating
pitch classes. Written responses consistent with these routes motivate a closer audit of
criterion~C3, but do not establish which computation the model performed. High unaided accuracy
alone is not evidence that the stated rule was unnecessary.

By question form, \QComparativePercept{}\% of \QComparativeN{} comparative items are first
solved with percept text, compared with \QAbsolutePercept{}\% of \QAbsoluteN{} absolute items.
Categorical items have the largest unaided success share, \QCategoricalCapable{}\% of
\QCategoricalN{}. These differences are descriptive: question form, source, and perceptual
axis are not independently varied. Further controlled comparisons are needed to attribute them
to perceptual difficulty or recognition shortcuts.

\subsection{Validation still required}\label{sec:pendingvalidation}
The current evaluation does not include a human listening study. Such a study is needed to
test whether the cues are audible, whether percept text is accurate, and whether the supplied
rule and observations make the questions answerable. Repeated runs, option permutations and
percept paraphrases are also needed to assess the stability of the operational labels.
The knowledge-recall probe also requires independent grading and repeated measurements before
its labels can support stronger conclusions. Finally, additional sources
per perceptual axis and a separate contamination set are needed for cross-source and
leak-detection claims.

\section{Limitations}\label{sec:limitations}
\begin{itemize}
  \item \textbf{The percept must be verbalizable.} The protocol can only diagnose reasoning whose
        required observation can be written in a sentence. Percepts that resist short textual
        description, such as timbre, texture or simultaneous streams, are out of scope by
        construction, and this excludes much of what audio perception involves.
  \item \textbf{Attribution is behavioural.} ``Failed perception'' means an item becomes solvable
        once the percept is supplied as text, not that an identified component of the model failed
        to encode it.
  \item \textbf{Scale and balance.} The build is still in progress: several families are small,
        so per-family intervals are wide, and the family sizes are not yet balanced against each
        other. \NSynth{} of \NItems{} items are a controlled transformation of a real recording
        rather than an unmodified one. None is synthesized from tones or noise alone; that
        material was removed from the build.
  \item \textbf{Transformed recordings.} Controlled transformations provide known parameters,
        but may make the stated physical relation easier to apply than in an uncontrolled
        recording. Performance on these items does not establish the validity of the relation
        across real acoustic environments.
  \item \textbf{Selection effects.} Items were selected using the construction checks, including
        agreement between extracted percepts and reference values. Results therefore describe
        this selected set, rather than a random sample of acoustic reasoning tasks.

  \item \textbf{Rule-domain and percept skew.} Physics accounts for more than half the items, with
        music and counting next and phonetics, clinical and signal reasoning represented by a
        single family each (Table~\ref{tab:facets}). Frequency is over-represented among the
        percepts for the same reason: a quantity that shows up as a pitch, a shift in pitch or a
        rate is the kind most often paired with an instrument that can measure it independently.
        Percepts that turn on level, timing or bandwidth are a minority, and the benchmark should
        not be read as covering audio perception evenly.
  \item \textbf{Percept validation.} Percept text is produced by our measurement routines.
        The stored checks cover \NChkRuleCovered{} of \NItems{} items, with disagreements
        remaining to be resolved. A systematic human listening study has not been completed.
  \item \textbf{Evaluation scope.} Only one model has complete reported results. The probe
        uses that same model as grader, and label stability has not been tested across repeated
        runs, option permutations, or percept paraphrases.
  \item \textbf{Release status.} Data and code packaging is proposed. Source permissions and
        final release licences require confirmation, including for transformed audio.

\end{itemize}

\section{Conclusion}
AuDiR evaluates audio reasoning through explicit rule and percept interventions. The current
\NItems{}-item build supports inspection of both the supplied information and each model's
response pattern. In the preliminary evaluation of \MAName{}, supplying percept text produces
a much larger accuracy gain than supplying the rule alone. No-audio accuracy is consistent with
chance in aggregate. These findings motivate further study of how acoustic observations are
used during reasoning, while incomplete percept validation, possible recognition shortcuts and
non-monotonic responses constrain stronger attribution. Human validation, additional models and
robustness checks are needed before the profiles can reliably guide model improvements.

\bibliographystyle{IEEEtran}
\bibliography{refs}

\end{document}